\documentclass[11pt]{article}

\usepackage[T1]{fontenc}
\usepackage[utf8]{inputenc}
\usepackage[margin=1in]{geometry}
\usepackage{amsmath,amssymb}
\usepackage{caption}
\usepackage{graphicx}
\usepackage{hyperref}
\usepackage[section]{placeins}
\usepackage{url}
\usepackage{xcolor}

\title{A Note on TurboQuant and the Earlier DRIVE/EDEN Line~of~Work}
\author{\small
\begin{tabular}{@{}c@{\hspace{0.8em}}c@{\hspace{0.8em}}c@{}}
Ran Ben-Basat & Yaniv Ben-Itzhak & Gal Mendelson \\
UCL and Broadcom & VMware Research by Broadcom & North Carolina State University \\[1em]
Michael Mitzenmacher & Amit Portnoy & Shay Vargaftik \\
Harvard University & Microsoft & VMware Research by Broadcom
\end{tabular}
}
\date{}

\begin{document}

\maketitle

\begin{abstract}
This note clarifies the relationship between the recent TurboQuant work and the earlier DRIVE (NeurIPS 2021) and EDEN (ICML 2022) schemes. DRIVE is a 1-bit quantizer that EDEN extended to any $b>0$ bits per coordinate; we refer to them collectively as EDEN.

First, {\sc TurboQuant$_{\text{mse}}$} is a special case of EDEN obtained by fixing EDEN's scalar scale parameter to $S=1$. EDEN supports both biased and unbiased quantization, each optimized by a different $S$ (chosen via methods described in the EDEN works). The fixed choice $S=1$ used by TurboQuant is generally suboptimal, although the optimal $S$ for biased EDEN converges to $1$ as the dimension grows; accordingly {\sc TurboQuant$_{\text{mse}}$} approaches EDEN's \mbox{behavior for large $d$}.

Second, {\sc TurboQuant$_{\text{prod}}$} combines a biased $(b-1)$-bit EDEN step with an unbiased 1-bit QJL quantization of the residual. It is suboptimal in three ways: (1) its $(b-1)$-bit step uses the suboptimal $S=1$; (2) its 1-bit unbiased residual quantization has worse MSE than (unbiased) 1-bit EDEN; (3) chaining a biased $(b-1)$-bit step with a 1-bit unbiased residual step is inferior to unbiasedly quantizing the input directly with $b$-bit EDEN.

Third, some of the analysis in the TurboQuant work mirrors that of the EDEN works: both exploit the connection between random rotations and the shifted Beta distribution, use the Lloyd-Max algorithm, and note that Randomized Hadamard Transforms can replace uniform random rotations.

Experiments support these claims: \textbf{biased} EDEN (with optimized $S$) is more accurate than {\sc TurboQuant$_{\text{mse}}$}, and \textbf{unbiased} EDEN is markedly more accurate than {\sc TurboQuant$_{\text{prod}}$}, often by more than a bit (e.g., 2-bit EDEN beats 3-bit {\sc TurboQuant$_{\text{prod}}$}). We also repeat all accuracy experiments from the TurboQuant paper, showing that EDEN outperforms it in every setup we have tried.
\end{abstract}

\section{Introduction}

On March 24, 2026, Google publicly highlighted TurboQuant~\cite{turboquant_iclr}, their recently accepted
ICLR 2026 paper, as a breakthrough in AI memory efficiency in an official blog post
\cite{google_turboquant_blog,turboquant_iclr}.
That public framing quickly spilled into financial coverage.
Investing.com, carrying Reuters credit, reported on March 25, 2026 that Samsung
Electronics fell 4.8\% and SK Hynix 5.9\%, while U.S.-listed memory peers Micron,
SanDisk, Western Digital, and Seagate fell between 3\% and 6\% \cite{investing_turboquant}.
Seoul Economic Daily likewise covered the March 26--27, 2026 selloff and linked it to
concerns that lower AI memory requirements could reduce future demand for advanced
memory chips, while also reporting the counterargument that cheaper AI may expand overall
demand over time \cite{sedaily_turboquant,sedaily_turboquant_followup}.

However, as we explain in this note, the {\sc TurboQuant$_{\texttt{mse}}$} algorithm is a suboptimal special case of the biased variant of the EDEN\footnote{EDEN was also contributed to Intel's OpenFL~\cite{openfl_eden_pipeline_code,vmware_eden_openfl_2022} in 2022.} algorithm from ICML 2022, and the unbiased variant of EDEN has better accuracy than that of the unbiased variant {\sc TurboQuant$_{\text{prod}}$}. We also show that much of the analysis used in TurboQuant previously appeared in the DRIVE~\cite{drive} and EDEN~\cite{eden} papers (which we collectively refer to as EDEN) and conduct experiments to empirically compare the algorithms.\footnote{We note that the authors of the RaBitQ~\cite{gao2024rabitq} paper have expressed similar concerns (e.g., \cite{gao2026turboquant}) regarding their paper; EDEN and DRIVE also predate the RaBitQ work, which we have recently communicated to the RaBitQ authors. Here, we focus on comparing TurboQuant and EDEN.}
We hope that demonstrating the advantages of using EDEN will support its continued adoption in emerging systems more rapidly.

\section{Preliminaries}

We use $x \in \mathbb{R}^d$ for the input vector, and we write $Q$ and $Q^{-1}$ for the
quantization and dequantization maps.
Following EDEN, we also use $
\eta_x := \frac{\sqrt{d}}{\|x\|_2},
$ 
so that $\eta_x R(x)$ has coordinates on the standard-normal scale after rotation.
TurboQuant formulates the reconstruction objective through the mean-squared distortion
$D_{\mathrm{mse}} := \mathbb{E}\|x-Q^{-1}(Q(x))\|_2^2$ and the inner-product distortion
$D_{\mathrm{prod}} := \mathbb{E}|\langle y,x\rangle-\langle y,Q^{-1}(Q(x))\rangle|^2$
in, equations~(1)-(2)~\cite{turboquant_iclr}.
EDEN~\cite{eden} utilizes the vector-normalized mean-squared error
$\mathrm{vNMSE} := \mathbb{E}\|x-\hat{x}\|_2^2/\|x\|_2^2$.
These are the same normalization conventions we use throughout this note.

The common geometric setup is also the same.
After a uniform random rotation, the coordinates become identically distributed.
The DRIVE paper \cite[Lemma 8, Appendix A.4 of the supplemental material]{drive} further explains 
% explains this in Section~2.2.1 on p.~3 and notes in footnote~5 on p.~5 that the
that the exact distribution of each coordinate with finite $d$ is a shifted Beta distribution that rapidly approaches a normal distribution as $d$ increases
\cite{eden}.
TurboQuant finds the same coordinate distribution in Lemma~1 \cite{turboquant_iclr}.

\subsection*{Biased and Unbiased Scales in EDEN}

The key distinction for the present note is the choice of reconstruction scale.
EDEN's unbiased scale is introduced in Theorem~2.1: for
\[
S_{\mathrm{unb}}(x,R) = \frac{\|x\|_2^2}{\langle R(x),Q(\eta_x R(x))\rangle},
\]
EDEN proves $\mathbb{E}[\hat{x}] = x$ \cite{eden}.
EDEN then proves a corresponding vNMSE bound in Theorem~2.3 and its asymptotic form in Corollary~2.4.

One can alternatively choose the scale factor to minimize the distortion, without the unbiasedness constraint, by setting 
\[
S_{\mathrm{bias}}(x,R)=\frac{\langle R(x),Q(\eta_x R(x))\rangle}{\|Q(\eta_x R(x))\|_2^2}.
\]
With this setting, $\hat{x}=S_{\mathrm{bias}}(x,R)R^{-1}Q(\eta_x R(x))$ is the best scalar
rescaling of the chosen codeword in squared error.
For the one-bit precursor DRIVE, this dichotomy is explicit: Lemma~1 and Theorem~2 analyze the MSE-minimizing scale, while Theorem~3, Theorem~4, and Corollary~1 on analyze the unbiased scale and its distributed-mean-estimation consequences
\cite{drive}.
EDEN generalizes this picture to arbitrary bitwidths: Section~3 chooses the
Lloyd--Max quantizer that minimizes scalar MSE, while Section~2.3 and Corollary~2.4 explain how to combine the same scalar quantizer with the unbiased scale \cite{eden}.
This is exactly the distinction that matters for TurboQuant:
{\sc TurboQuant$_{\texttt{mse}}$} uses a fixed choice $S=1$ that is biased, whereas EDEN shows that an appropriate choice of scale leads to an unbiased result, and further shows how to choose a (different) optimal \mbox{$S$ for biased results.}

% \section{Background and Chronology}
% For the historical record, the relevant sequence is straightforward.
% DRIVE appeared in NeurIPS 2021 \cite{drive};
% EDEN appeared in ICML 2022 \cite{eden}; 
%  TurboQuant was first posted to arXiv on 2025 \cite{turboquant}, posted as a blog post on
% March 24, 2026 \cite{google_turboquant_blog}, and it was presented as an ICLR 2026
% poster on April 25, 2026 % \cite{turboquant_iclr}.

%This chronology matters because the right point of comparison depends on the bitwidth.
%DRIVE is the one-bit predecessor.
%EDEN is the general-$b$ predecessor.
%Therefore, when discussing algorithms or empirical results for arbitrary bitwidths
%$b \in \{1,2,3,\ldots\}$, the correct earlier reference point is EDEN, with DRIVE
%serving mainly as the $b=1$ ancestor of that broader framework.
%
%
%
%We also note for the record that, before preparing this public note, we reached out
%privately to the Google TurboQuant authors with these
%observations and did not receive a reply.\footnote{This sentence concerns private correspondence rather than material appearing in either paper, so it does not carry a paper citation.}

% \section{\texorpdfstring{{\sc TurboQuant$_{\texttt{mse}}$} as EDEN with $S=1$}{TurboQuant-MSE as EDEN with S=1}}\label{sec:tqmse}

\section{\texorpdfstring{{\sc TurboQuant$_{\texttt{mse}}$} as EDEN with $S=1$}{TurboQuant-MSE as EDEN with S=1}}\label{sec:tqmse}

Our first observation concerns the MSE-oriented TurboQuant construction.
Viewed through the EDEN parametrization, {\sc TurboQuant$_{\texttt{mse}}$} corresponds to the
special case obtained by fixing the EDEN scale parameter to $S=1$.
% In that sense, it lies inside the EDEN family rather than outside it.

To make this relationship explicit, Figure~\ref{fig:pseudocode} shows a unified  pseudocode for {\sc TurboQuant$_{\texttt{mse}}$} and (both biased and unbiased) EDEN.
Relative to the notation of Section~2, the figure writes the dequantized rotated
codeword in the same scale as the rotated vector, namely
\[
q := Q(\eta_x y)/\eta_x.
\]
In this normalization, the common structure is especially transparent:
rotate, quantize coordinates, apply the inverse rotation, and reconstruct.
The three methods differ only in their choice of the final reconstruction
scale $S$: {\sc TurboQuant$_{\texttt{mse}}$} fixes $S=1$, EDEN-biased uses the
MSE-minimizing scalar, and EDEN-unbiased uses the unbiased scalar from
Figure~1 of \cite{eden}.

The scalar quantizers themselves also coincide.
For the one-bit case, the same two-point reconstruction is already explicit in DRIVE:
Algorithm~1 reconstructs with values in $\{\pm S\}$, and Lemma~1 gives the
biased MSE-minimizing scale explicitly as
$
S_{\mathrm{bias}}=\frac{\|R(x)\|_1}{d},
$
so the one-bit DRIVE reconstruction levels are exactly
$\{\pm \|R(x)\|_1/d\}$ \cite[Lemma~1]{drive}.
%Theorem~2 then gives the resulting vNMSE exactly as
%$(1-2/\pi)(1-1/d)$, which tends to $1-2/\pi \approx 0.363$.
EDEN Section~3, specifically Example~1 and Example~2, gives the standard-normal
Lloyd--Max codebooks
$QI_1 \approx \{\pm 0.79788\}$ and
$QI_2 \approx \{\pm 0.45278,\pm 1.51042\}$ \cite{eden}.
 TurboQuant (Section~3.1 and Algorithm~1) lists the same centroids, up to the
paper's $\sqrt{d}$ normalization.
For $b=1$ it uses $\pm \sqrt{2/\pi}/\sqrt{d}$, and for $b=2$ it uses
$\{\pm 0.453/\sqrt{d},\pm 1.51/\sqrt{d}\}$ \cite{turboquant_iclr}.
%Likewise,  TurboQuant Theorem~1 reports
%$D_{\mathrm{mse}} \approx 0.36, 0.117, 0.03, 0.009$ for $b=1,2,3,4$ \cite{turboquant}.
%DRIVE explicitly gives the $1-2/\pi\approx 0.363$ as the vNMSE for when $b = 1$ (Theorem 2).
%Combining EDEN's Corollary~2.4 with the same Lloyd--Max quantizers of
%Section~3 yields exactly these biased/MSE constants for $b=2,3,4$.  
%So both the centroids and the resulting MSE values are the same.

Further to these similarities, we note that the analysis of EDEN provides tighter bounds. For example, for $b=1$, DRIVE shows~\cite[Theorem~2]{drive} that the vNMSE is \textit{exactly} $(1-2/\pi)(1-1/d)$, which is bounded by $1-2/\pi \approx 0.363$. 
TurboQuant proved~\cite[Theorem~1]{turboquant_iclr} that $D_{\mathrm{mse}}\le \frac{\sqrt 3 \pi}{2}\cdot \frac{1}{4^b}$, which gives $\sqrt 3 \pi / 8 \approx 0.68$ for $b=1$. The authors mention~\cite[Theorem~1]{turboquant_iclr} that for $b=1$, $D_{\mathrm{mse}}\approx 0.36$, but this is not proven and seems to be based on empirical observation.  

%Nor is the underlying coordinate-law analysis new.
As mentioned, these results are derived from the fact that the coordinate distribution after rotation, whose analysis also appears in the DRIVE paper~\cite[Lemma 8, Appendix A.4 of the supplemental material]{drive}, follows a shifted Beta distribution (which converges to a normal distribution as the dimension grows large).  

\begin{figure}[!htbp]
    \centering
    \fbox{%
    \begin{minipage}[t]{0.93\linewidth}
    \small
    \textbf{{\sc TurboQuant$_{\texttt{mse}}$}/EDEN pseudocode }\par
    \medskip
    \textbf{Setup}\par
    1. Generate the shared random rotation matrix $\Pi$.\par
    2. Construct the Lloyd–Max centroid codebook $c_1,\dots,c_{2^b}$ for the rotated coordinates.\par
    \medskip
    \textbf{Quantize}\par
    3. Compute $y \gets \Pi x$.\par
    4. For each $j \in [d]$, set
    $\mathrm{idx}_j \gets \arg\min_{k \in [2^b]} |y_j-c_k|$.\par
    5. Set $q_j \gets c_{\mathrm{idx}_j}$ for each $j \in [d]$.\par
    6. Choose the reconstruction scale $S$.\par
    7. Send $(q,S)$.\par
    \medskip
    \textbf{Dequantize}\par
    8. Regenerate $\Pi$ from the same seed.\par
    9. Output $\hat{x} \gets S \Pi^\top q$.\par
    \medskip
    \textbf{Relevant choices of $S$}\par
    10. {\sc TurboQuant$_{\texttt{mse}}$}: $S \gets 1$.\par
    11. EDEN-biased: $S \gets \langle y,q \rangle / \lVert q \rVert_2^2$.\par
    12. EDEN-unbiased: $S \gets \lVert x \rVert_2^2 / \langle y,q \rangle$.\par
    \end{minipage}}
    \caption{Unified {\sc TurboQuant$_{\texttt{mse}}$}/EDEN pseudocode.
    This presentation is written in the rotated-codeword scale
    $q=Q(\eta_x y)/\eta_x$, so the difference between the three methods is
    entirely in the reconstruction factor $S$.
    Fixing $S=1$ recovers {\sc TurboQuant$_{\texttt{mse}}$}, while the two
     choices of $S$ recover EDEN-biased and EDEN-unbiased.}
    \label{fig:pseudocode}
\end{figure}

\begin{figure}[!htbp]
    \centering
    \includegraphics[width=\textwidth]{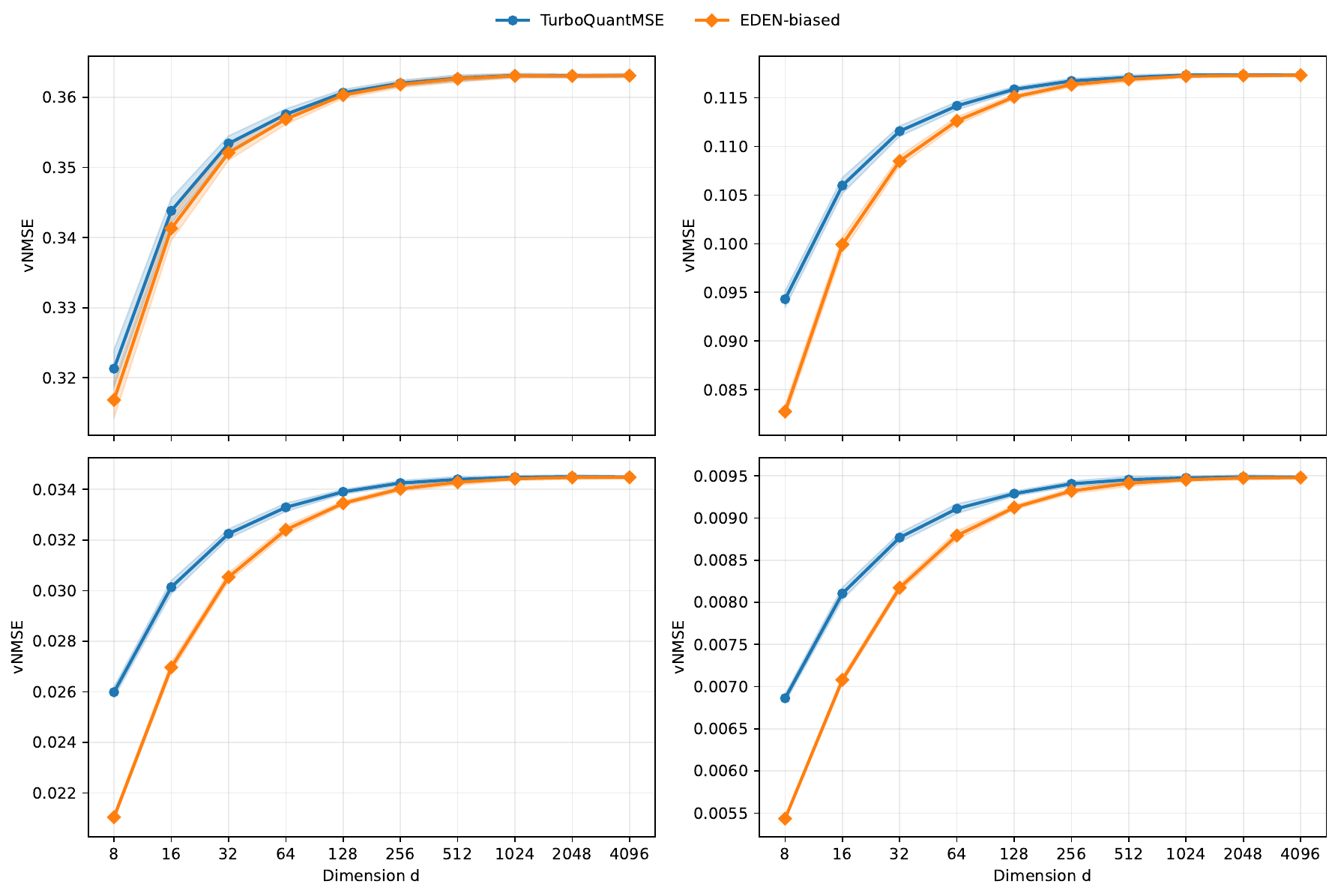}
    \caption{Comparison between {\sc TurboQuant$_{\texttt{mse}}$} and EDEN-biased, i.e.\ EDEN
    with the MSE-optimized scale, shown as a function of dimension for bitwidths
    $b \in \{1,2,3,4\}$.
    The four panels correspond, from left to right, to $b=1,2,3,4$.}
    \label{fig:mse}
\end{figure}

\begin{figure}[!htbp]
    \centering
    \begin{minipage}[t]{0.49\textwidth}
        \centering
        \includegraphics[width=\linewidth]{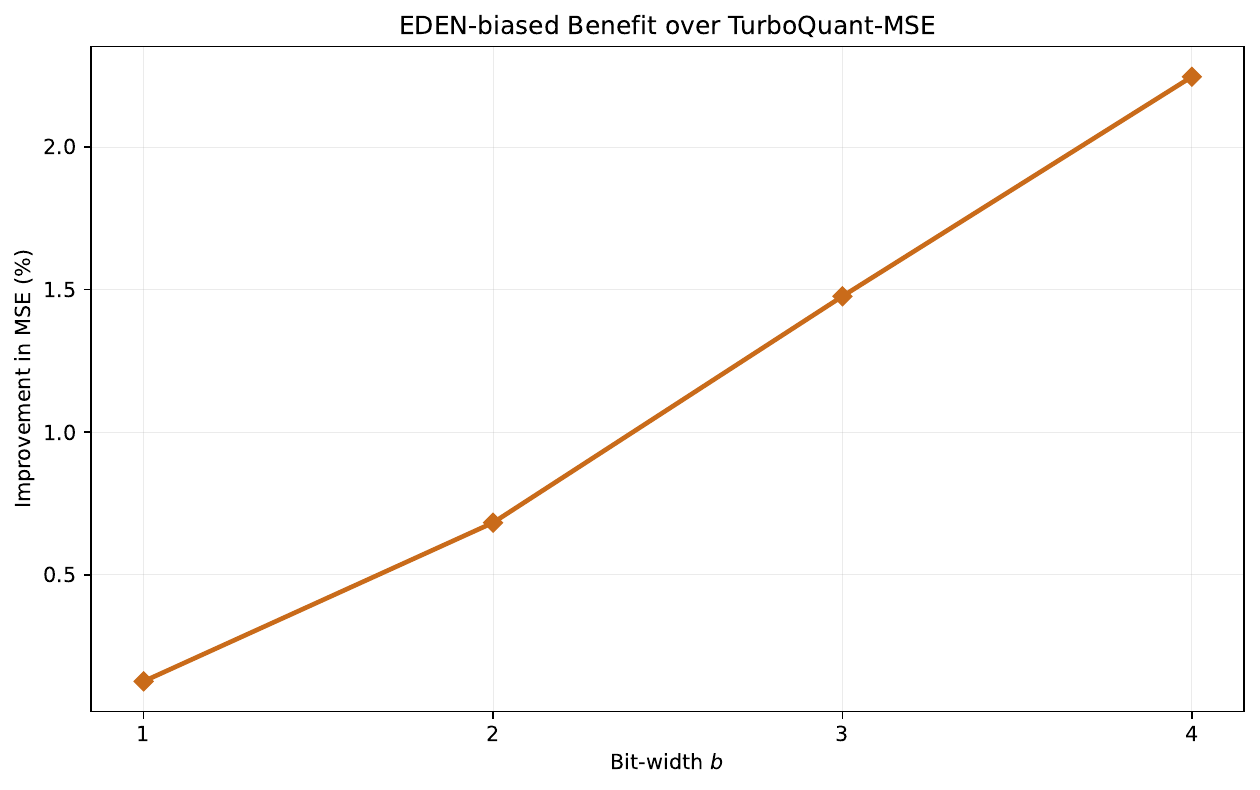}
        \captionof{figure}{Percentage MSE improvement of EDEN-biased over
        {\sc TurboQuant$_{\texttt{mse}}$} on \texttt{LogNormal}(0,1) vectors at fixed
        dimension $d=128$, for bitwidths $b=1,2,3,4$.
        The relative gain is $0.13\%$, $0.68\%$, $1.48\%$, and $2.25\%$,
        respectively.}
        \label{fig:biased_gap}
    \end{minipage}\hfill
    \begin{minipage}[t]{0.49\textwidth}
        \centering
        \includegraphics[width=\linewidth]{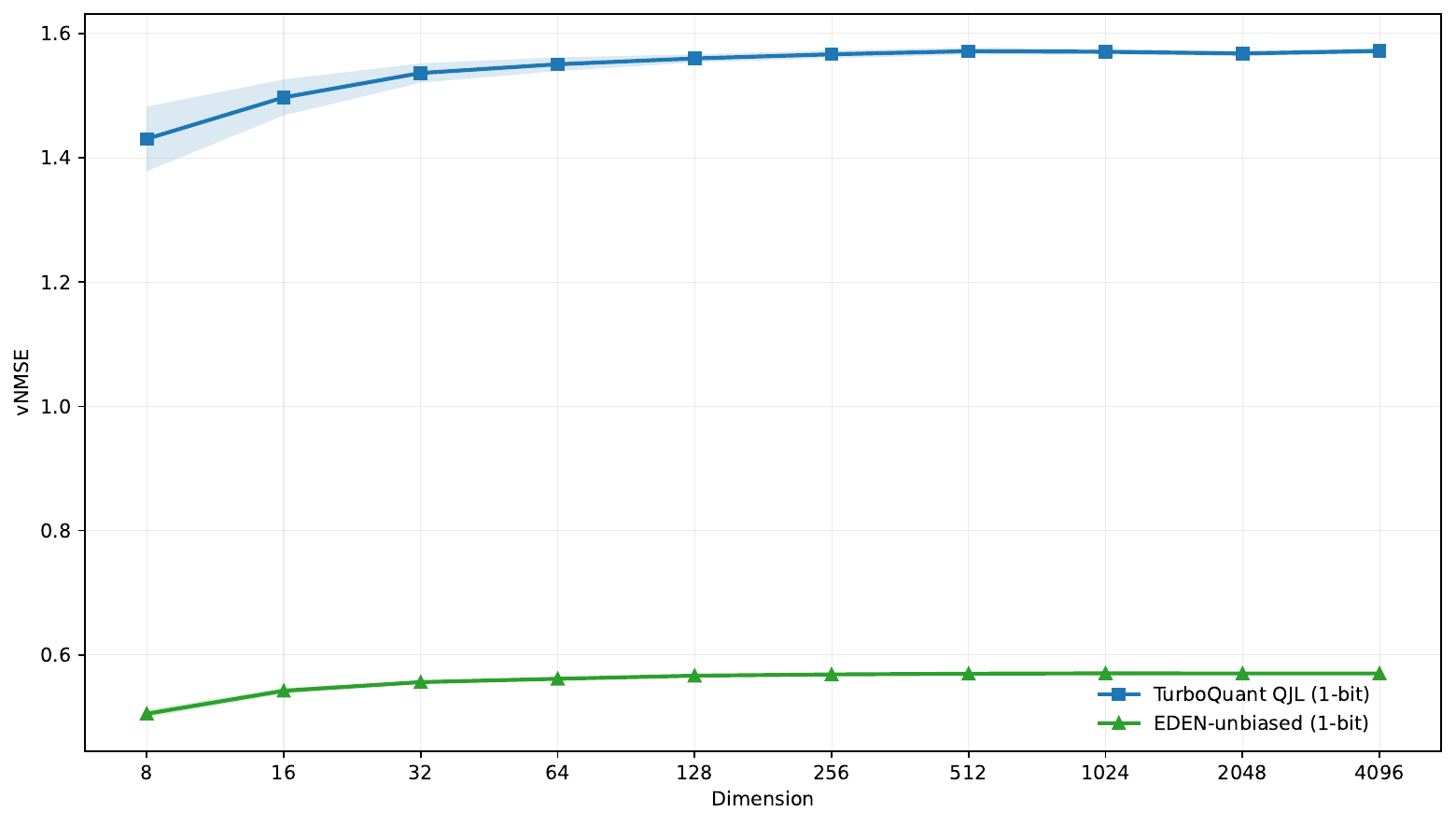}
        \captionof{figure}{Comparison between the pure one-bit  TurboQuant QJL estimator and
        \texttt{EDEN-unbiased}, shown as a function of dimension.
        This figure isolates only the one-bit residual-style stage used by
        {\sc TurboQuant$_{\text{prod}}$}.
        Throughout the plotted range, the QJL estimator is substantially weaker than
        the earlier unbiased DRIVE (e.g., 1-bit EDEN-unbiased) quantizer.}
        \label{fig:qjl}
    \end{minipage}
\end{figure}

Correspondingly, once Figure~\ref{fig:pseudocode} is written in this unified form,
the empirical behavior is exactly what one should expect.
Fixing $S=1$ in {\sc TurboQuant$_{\texttt{mse}}$} performs less well than the
biased version of EDEN (hereafter referred to as EDEN-biased), which chooses
the scale to minimize the resulting distortion.
% Consequently, {\sc TurboQuant$_{\texttt{mse}}$} is a restricted instantiation of the more general
% EDEN design space, and there is no reason to expect it to match the optimized EDEN
% choice.
Figure~\ref{fig:mse} shows this comparison across dimensions and bitwidths.
Figure~\ref{fig:biased_gap} further shows the accuracy gap across bitwidth for a specific $d=128$ dimension.

For Figures~\ref{fig:mse}, \ref{fig:qjl}, and \ref{fig:prod}, each plotted point
is the mean over paired repetitions using the same lognormal sample seed and the
same quantizer seed across the compared methods.\footnote{We note that the actual input distribution is irrelevant here as the algorithms randomly rotate the input.}
The displayed $95\%$ confidence intervals are
$1.96\,s/\sqrt{n}$, where $s$ is the sample standard deviation of the per-pair
metric and $n$ is the number of paired seeds for that dimension.
In these sweeps, $n=256$ for $d \le 128$, $n=128$ for $d=256$, $n=64$ for
$d \in \{512,1024\}$, $n=32$ for $d=2048$, and $n=16$ for $d=4096$.

Across all plotted bitwidths and dimensions, the EDEN-biased curve lies below the
{\sc TurboQuant$_{\texttt{mse}}$} curve.
The gap is most visible in lower dimensions, but it remains throughout the
displayed range.  In large dimensions, for EDEN-biased $S$ does converge to 1, and correspondingly we see {\sc TurboQuant$_{\texttt{mse}}$} does approach EDEN-biased performance for larger dimension.
% Thus this dimension-sweep comparison supports the simple interpretation above:
% {\sc TurboQuant$_{\texttt{mse}}$} is an EDEN special case, and that special case is empirically
% suboptimal relative to EDEN with the proper scale choice.

\section{Unbiased EDEN has better accuracy than \texorpdfstring{{\sc TurboQuant$_{\texttt{prod}}$} }{TurboQuant-MSE as EDEN with S=1}}

We now consider the relationship of {\sc TurboQuant$_{\text{prod}}$} and EDEN.

TurboQuant$_{\text{prod}}$ has two logical steps: It first uses TurboQuant$_{\text{mse}}$ with $b-1$ bits to quantize the input. Then, it calculates the error (the residual vector) and quantizes it with one bit per coordinate using the Quantized Johnson Lindenstrauss (QJL) method \cite{qjl}.

We find the unbiased variation of EDEN (EDEN-unbiased) outperforms TurboQuant$_{\text{prod}}$.
Upon examination, we find TurboQuant$_{\text{prod}}$ is suboptimal in several ways.
\begin{enumerate}
\item Its first stage uses only $(b-1)$ bits, and that stage is itself just the
      $S=1$ special case of EDEN rather than the EDEN-biased choice that minimizes
      MSE for the same codeword.
\item Its second stage uses a one-bit QJL quantization of the residual, and this
      one-bit estimator is provably and empirically much weaker than the earlier one-bit
      unbiased DRIVE quantizer (vNMSE for large dimension converges to approximately 0.571 for DRIVE vs 1.57 for QJL).
\item In fact, splitting into a biased quantization with $b-1$ bits followed by a $1$-bit unbiased quantization is less accurate than using all $b$ bits for unbiased quantization with EDEN.
\end{enumerate}

\begin{figure}[!htbp]
    \centering
    \includegraphics[width=.84279\textwidth]{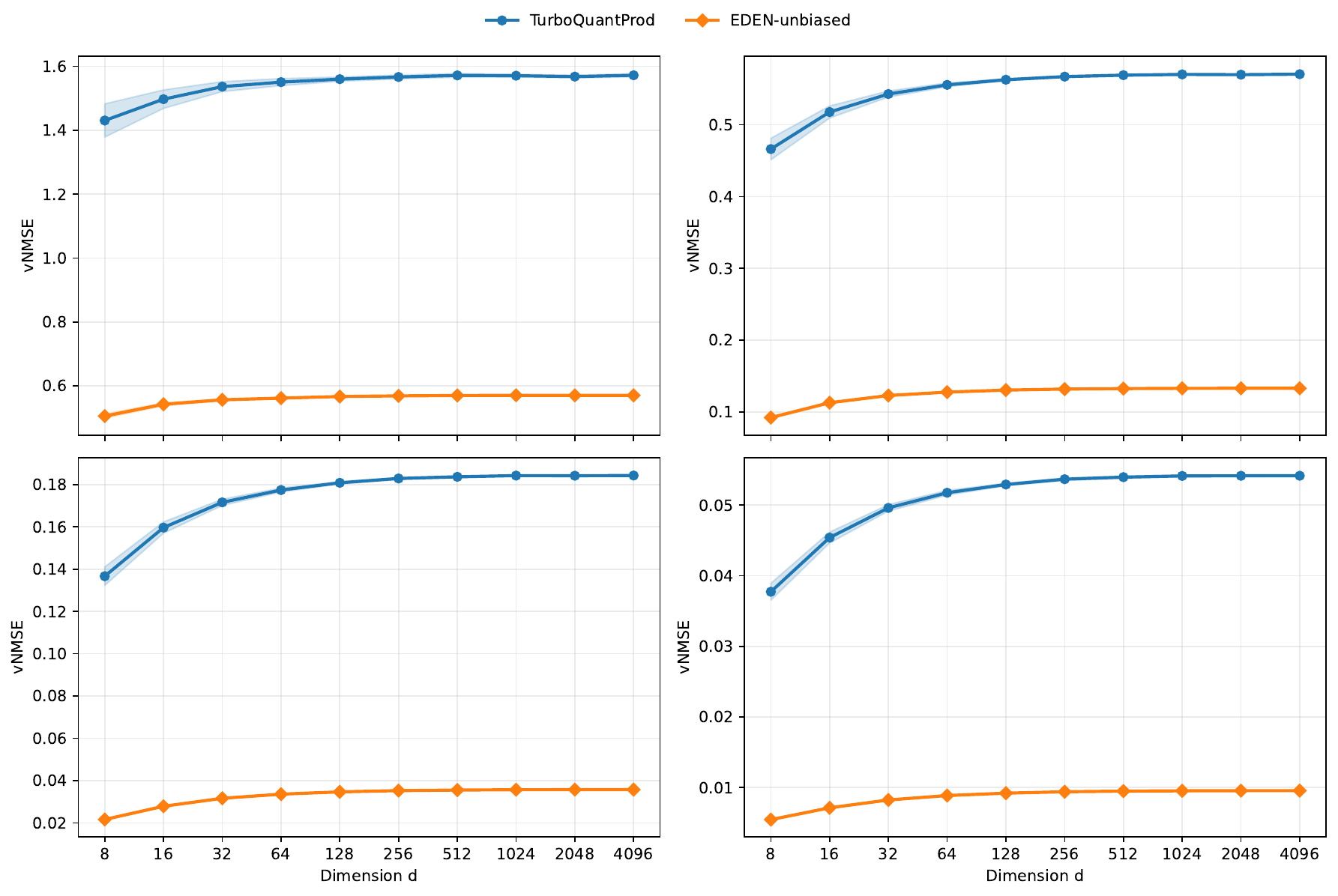}
    \caption{Comparison between {\sc TurboQuant$_{\text{prod}}$} and the unbiased EDEN
    baseline, shown as a function of dimension for bitwidths $b \in \{1,2,3,4\}$.
    The four panels correspond, from left to right, to $b=1,2,3,4$.
}
    \label{fig:prod}
\end{figure}

Figure~\ref{fig:prod} compares EDEN-unbiased and TurboQuant$_{\text{prod}}$ across dimensions and
bitwidths.
Throughout the plotted range, {\sc TurboQuant$_{\text{prod}}$}
exhibits substantially larger error than the unbiased EDEN baseline.
The separation is large for every shown bitwidth, and it remains large as the
dimension increases.
Notice that the gap is often worth more than a whole bit (e.g., EDEN with 2 bits is more accurate than {\sc TurboQuant$_{\text{prod}}$} with 3 bits, and with 3 bits EDEN is better than {\sc TurboQuant$_{\text{prod}}$} with 4 bits).

We also note that even if one decides to split the quantization to biased $b-1$ bits followed by unbiased $1$-bit quantization, it is better to use DRIVE than QJL for this purpose.
This is visible directly in Figure~\ref{fig:qjl}, which isolates the
one-bit residual-style estimator.
There, the pure QJL estimator used by {\sc TurboQuant$_{\text{prod}}$} has much larger vNMSE
than \texttt{DRIVE-unbiased (i.e., 1-bit EDEN-unbiased)} across the entire displayed range.
So even before considering the first-stage split, the one-bit residual mechanism
is already a weak choice relative to the earlier \mbox{one-bit unbiased baseline.}

\section{Reproducing the  TurboQuant Paper's Empirical Behavior}

A contribution of the TurboQuant paper is to show that this compression mechanism can be applied in a variety of settings, such as nearest-neighbor queries using a mechanism based on inner products. We have also reproduced the accuracy experiments reported in the TurboQuant paper using their open-source code, while adding EDEN in order to make a comparison.\footnote{We did not compare the runtimes, but we expect them to be very similar.}

\begin{figure}[!htbp]
    \centering
    \includegraphics[width=\textwidth]{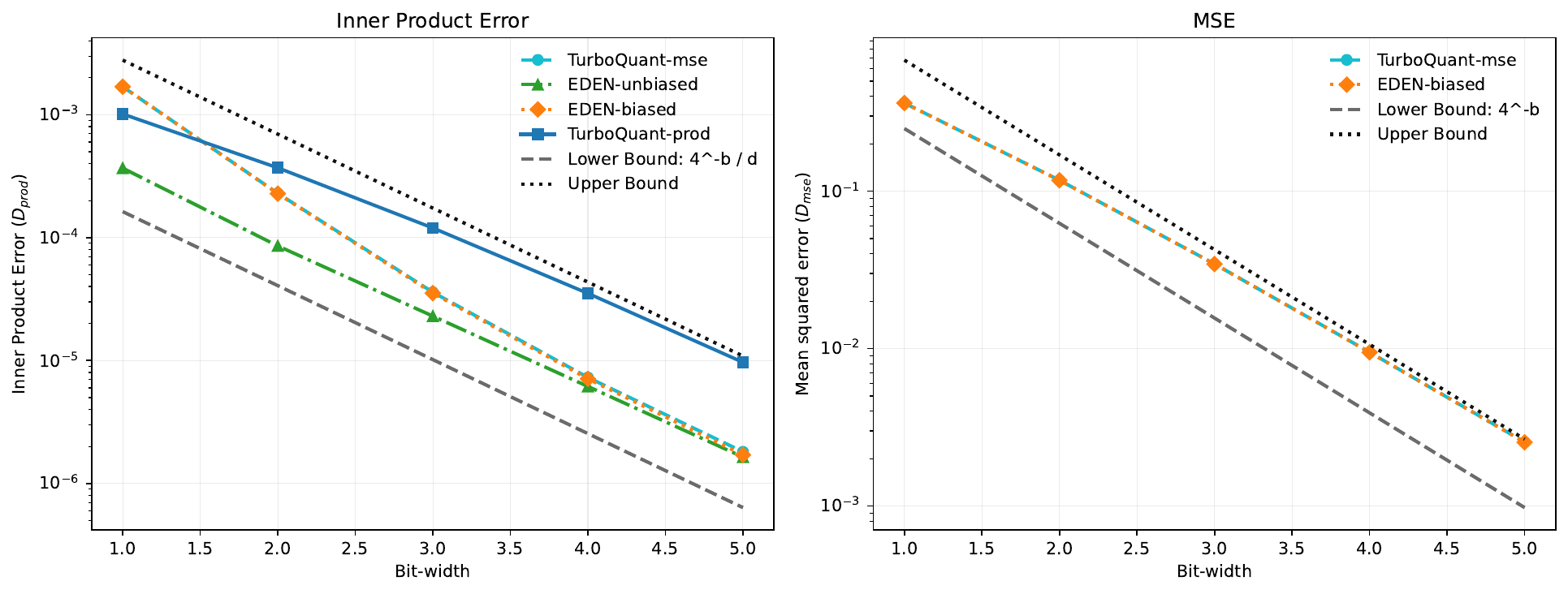}
    \caption{Aggregate accuracy curves in the reproduced  TurboQuant-style presentation.
    The right panel compares {\sc TurboQuant$_{\texttt{mse}}$} with EDEN-biased for MSE, while
    the left panel compares the inner-product-oriented methods as a function of
    bitwidth.
    In both panels, the horizontal axis is the bitwidth $b$.}
    \label{fig:accuracy}
\end{figure}

Figure~\ref{fig:accuracy} shows that the error of EDEN-unbiased is markedly lower for inner product estimation than both {\sc TurboQuant$_{\text{mse}}$} and {\sc TurboQuant$_{\text{prod}}$}. It also shows that the MSE of EDEN-biased is comparable to but lower than TurboQuant$_{\text{mse}}$, consistent with our results from Section~\ref{sec:tqmse}.

\begin{figure}[!htbp]
    \centering
    \includegraphics[width=\textwidth, trim=0pt 0pt 0pt 1cm, clip]{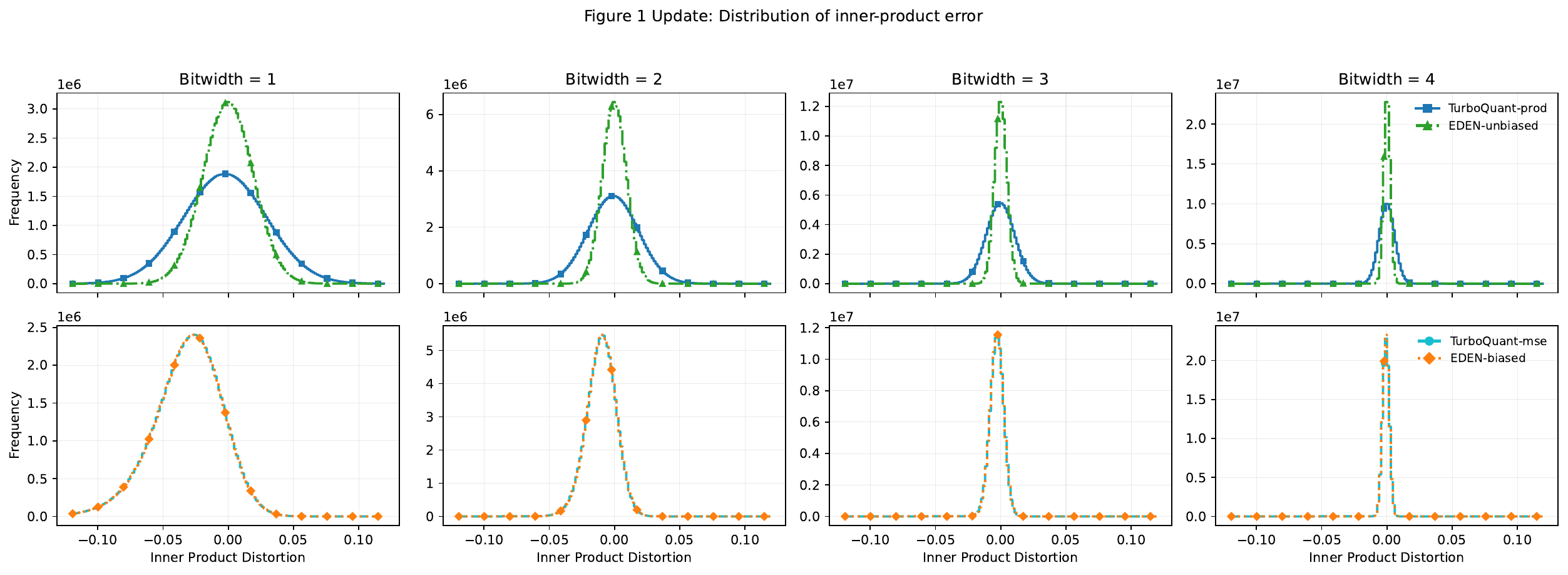}
    \caption{Distribution of inner-product error in the reproduced  TurboQuant-style
    presentation.
    The top row compares {\sc TurboQuant$_{\text{prod}}$} with EDEN-unbiased, and the bottom
    row compares {\sc TurboQuant$_{\texttt{mse}}$} with EDEN-biased, for bitwidths
    $b \in \{1,2,3,4\}$.
    Within each row, the four columns correspond from left to right to
    $b=1,2,3,4$.}
    \label{fig:error_distribution}
\end{figure}

Figure~\ref{fig:error_distribution} shows the full distributions of inner-product error. Once again, we observe the same patterns:  EDEN-unbiased outperforms {\sc TurboQuant$_{\text{prod}}$} and EDEN-biased is similar to but better than {\sc TurboQuant$_{\text{mse}}$}.

%At fixed dimension $d=128$, the same MSE-side advantage is visible in a simpler
%bitwidth summary.
%Figure~\ref{fig:biased_gap} plots the percentage improvement of EDEN-biased over
%{\sc TurboQuant$_{\texttt{mse}}$}.
%The gain is already positive at $b=1$ and then grows monotonically through
%$b=2,3,4$.

\begin{figure}[!htbp]
    \centering
    \includegraphics[width=\textwidth, trim=0pt 0pt 0pt 1cm, clip]{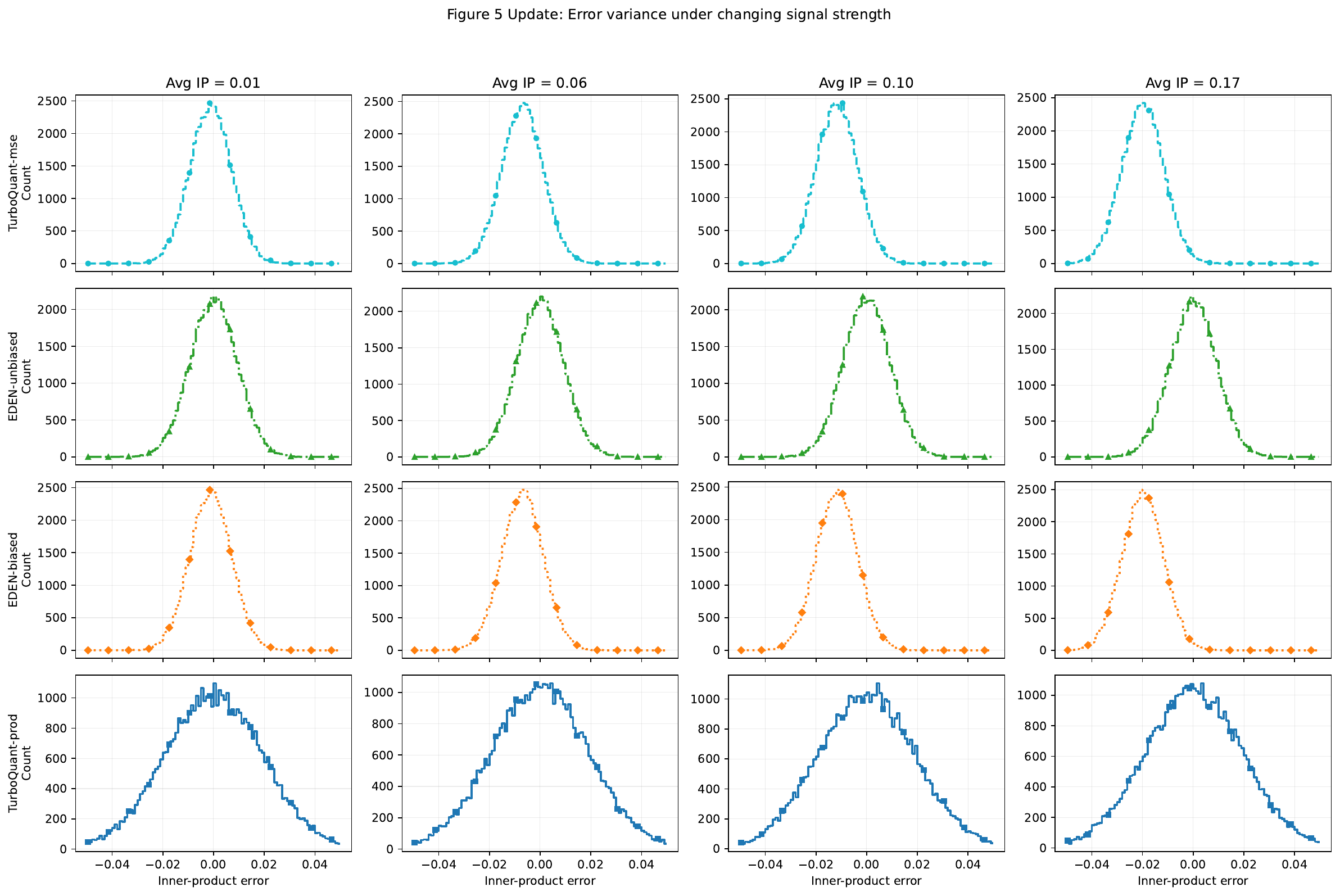}
    \caption{Inner-product error variance under changing signal strength in the
    reproduced  TurboQuant-style presentation.
    Across the displayed average inner-product regimes, EDEN-unbiased remains much more
    tightly concentrated than {\sc TurboQuant$_{\text{prod}}$}.}
    \label{fig:variance}
\end{figure}

\begin{figure}[!htbp]
    \centering
    \includegraphics[width=\textwidth]{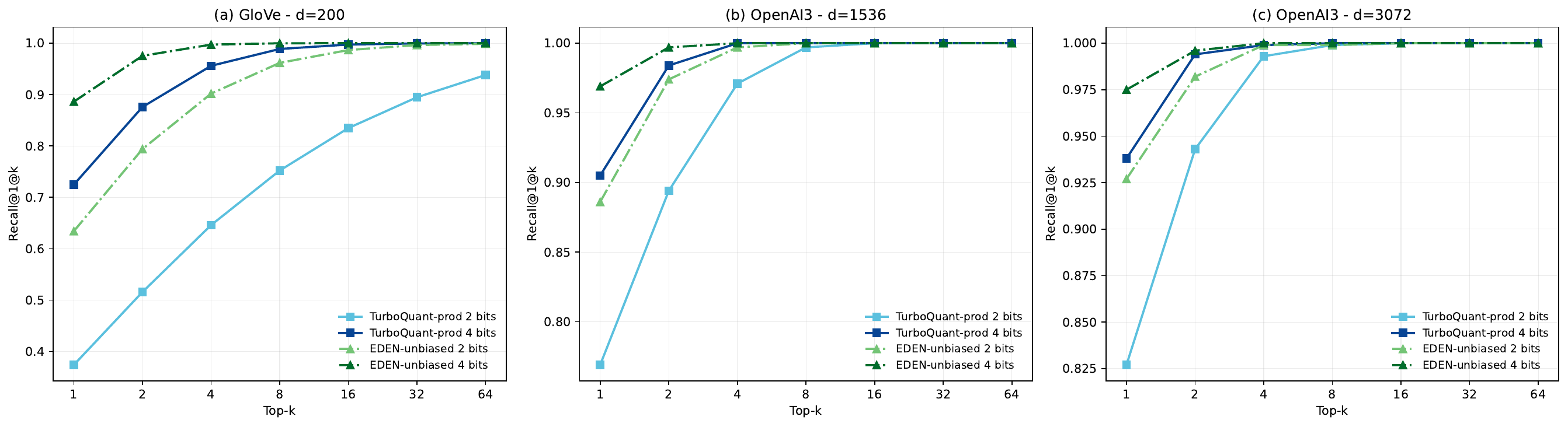}
    \caption{Downstream nearest-neighbor recall comparison in the reproduced
     TurboQuant-style presentation.
    EDEN-unbiased outperforms {\sc TurboQuant$_{\text{prod}}$} on the displayed GloVe and
    OpenAI3 settings at both two and four bits.}
    \label{fig:recall}
\end{figure}

Two additional reproduced figures reinforce the same conclusion in more specialized
settings.
Figure~\ref{fig:variance} shows that the inner-product error of
{\sc TurboQuant$_{\text{prod}}$} remains much more dispersed than that of EDEN-unbiased as the
average signal strength changes.
Figure~\ref{fig:recall} shows that the same qualitative gap persists in a downstream
nearest-neighbor retrieval task.

\section{Randomized Hadamard Transform}
We note that both DRIVE/EDEN and TurboQuant suggest using the Randomized Hadamard Transform (RHT) in practice instead of uniform random rotations, in order to reduce the computational cost of the rotation step.

We note that when using RHT, outputs from the RHT step may affect unbiasedness. However, both papers observed that it is essentially as accurate and nearly unbiased in practice.

This is not true for adversarial inputs (the DRIVE paper provides an example~\cite{drive}).  We note that our followup work QUIC-FL~\cite{basat2024quick} allows unbiased estimates with a single RHT, and that our recent results~\cite{2RHT} indicate that by using two consecutive RHTs, one can get provably nearly-unbiased results, in the sense that the bias vanishes polynomially in the dimension.

\section{Discussion}

% We note that both papers 
% There is also an important practical overlap regarding the rotation itself.
% TurboQuant states in Section~3.1 and Algorithm~1 that it uses a random
% rotation matrix $\Pi$ generated via QR decomposition \cite{turboquant}.
% EDEN had already pointed out the practical replacement of a full uniform random rotation by
% a randomized Hadamard transform in Section~5.1 on p.~8 and in Appendix~H / Figure~6 on
% pp.~26--27 \cite{eden}; the one-bit predecessor DRIVE likewise develops the structured
%Hadamard version explicitly in Section~6 on pp.~5--6 \cite{drive}.
%So the practical use of an RHT in place of a full uniform random rotation was already part
% of our earlier line of work.

The TurboQuant work suggests that compression techniques based on randomized rotations may have important potential applications to improve efficient use of AI memory.  This note has shown that some of the compression algorithms and much of the corresponding analysis for the TurboQuant methods appeared previously in the DRIVE and EDEN papers. In fact, the {\sc TurboQuant$_{\texttt{mse}}$} algorithm is a special case of the biased variant of the EDEN, and the unbiased variant of EDEN has better accuracy than that of the unbiased variant {\sc TurboQuant$_{\text{prod}}$}. We hope that demonstrating the advantages of using EDEN, with its scaling factor, will support its continued adoption in emerging systems more rapidly.

\bibliographystyle{plain}
\bibliography{refs}

\end{document}